%% file: main.tex
\documentclass[conference]{IEEEtran}
\usepackage{setspace}
\usepackage{amsmath}
\usepackage{acronym}
\usepackage{bbm}
\usepackage[dvips]{color}
\usepackage{comment}
\usepackage{todonotes}
\usepackage{epsf}
\usepackage{siunitx}
\usepackage{epsfig}
\usepackage{times}
\usepackage{epsfig}
\usepackage{graphicx}
\usepackage{bbold}
\usepackage{geometry}
\usepackage{mathrsfs}
\usepackage{amssymb}
\usepackage{pdfpages}
\usepackage{epstopdf}
\usepackage{tcolorbox,bbold}
\usepackage{algorithm,algorithmicx,algpseudocode}
\makeatletter
\newcommand{\algmargin}{\the\ALG@thistlm}
\makeatother
\algnewcommand{\parState}[1]{\State%
  \parbox[t]{\dimexpr\linewidth-\algmargin}{\strut #1\strut}}

\usepackage[nomain,acronym,shortcuts]{glossaries}
\newcommand*{\acro}[3][]{\newacronym[#1]{#2}{#2}{#3}}
\include{acronyms}

\usepackage{amssymb}
\usepackage{url}
\usepackage{dsfont}
\usepackage{lettrine} % \lettrine[findent=1pt]{{{R}}}{}
\usepackage{cite}
\usepackage{amsmath,epsfig,amssymb,algorithm,algpseudocode,amsthm,cite,url}
\IEEEoverridecommandlockouts
\usepackage{subcaption}
\allowdisplaybreaks
\usepackage{csquotes}

\usepackage{verbatim}
\usepackage[english]{babel}
\usepackage{amsmath,amssymb}

\usepackage[justification=centering]{caption}
\usepackage{verbatim}

\newtheorem{definition}{\bf Definition}

\IEEEoverridecommandlockouts
\begin{document}
	%\pagenumbering{gobble}% Remove page numbers (and reset to 1)
%	\clearpage
	%\maketitle
	%Title.
	% ------
	\title{Disentangling Learnable and Memorizable Data via Contrastive Learning for Semantic Communications \vspace{-0.35cm}}
		\author{Christina Chaccour, \emph{Graduate Student Member, IEEE,} and Walid Saad, \emph{Fellow, IEEE,}\\
			\IEEEauthorblockA{Wireless@VT, Bradley Department of Electrical and Computer Engineering, Virginia Tech, Arlington, VA USA. \vspace{-0.8cm}}
	\thanks{\noindent This research was supported by the U.S. National Science Foundation under Grants CNS-2007635.}
}
\maketitle
	%\author{Mehdi Naderi Soorki$^{1,2}$, Walid Saad$^{1}$, and Mehdi Bennis$^{2}$\vspace*{0.15cm}\\
	%\small {$^{1}$Wireless@VT, Bradley Department of Electrical and Computer Engineering, Virginia Tech, Blacksburg, VA, USA.}\\
	%\small {$^{2}$Centre for Wireless Communications, University of Oulu, Finland.}\\
	%\small {Emails: \{mehdin,walids\}@vt.edu, mehdi.bennis@oulu.fi}
	%\vspace{-0.5cm}
	%  \thanks{This research was supported by the U.S. National Science Foundation under Grants CNS-1836802 and CNS-1814477.}%
	%}

	%-----------------------------------------------------
%		\title{\vspace{-0.5cm}On the Reliability of Wireless Virtual Reality at Terahertz (THz) Frequencies\vspace{-0.4cm}}
%	\author{\IEEEauthorblockN{Christina Chaccour\IEEEauthorrefmark{1},
%			Ramy Amer\IEEEauthorrefmark{2},
%			Bo Zhou\IEEEauthorrefmark{1},
%			and Walid Saad\IEEEauthorrefmark{1}}
%		\IEEEauthorblockA{\IEEEauthorrefmark{1}Wireless@ VT, Bradly Department of Electrical and Computer Engineering, Virginia Tech, Blacksburg, VA USA,}
%		\IEEEauthorblockA{\IEEEauthorrefmark{2}CONNECT, Trinity College, University of Dublin, Ireland.}
%		\IEEEauthorblockA{Emails:\{christinac, ecebo, walids\}@vt.edu, ramyr@tcd.ie \vspace{-11mm}}
%		
%	}
	
%	\maketitle

%	\thispagestyle{empty}

\begin{abstract}
Achieving artificially intelligent-native wireless networks is necessary for the operation of future 6G applications such as the metaverse. Nonetheless, current communication schemes are, at heart, a mere reconstruction process that lacks reasoning. One key solution that enables evolving wireless communication to a human-like conversation is semantic communications. In this paper, a novel machine reasoning framework is proposed to pre-process and disentangle source data so as to make it \emph{semantic-ready}. In particular, a novel contrastive learning framework is proposed, whereby instance and cluster discrimination are performed on the data. These two tasks enable increasing the cohesiveness between data points mapping to semantically similar content elements and disentangling data points of semantically different content elements. Subsequently, the semantic deep clusters formed are ranked according to their level of confidence. Deep semantic clusters of highest confidence are considered \emph{learnable, semantic-rich} data, i.e., data that can be used to build a language in a semantic communications system. The least confident ones are considered, \emph{random, semantic-poor, and memorizable data} that must be transmitted classically. Our simulation results showcase the superiority of our contrastive learning approach in terms of semantic impact and minimalism. In fact, the length of the semantic representation achieved is minimized by $57.22\%$ compared to vanilla semantic communication systems, thus achieving minimalist semantic representations.    
\end{abstract}
\vspace{0cm}
\section{Introduction}\label{sec:Intro}
To enable emerging services like the metaverse, digital twins, Web 3.0, and Industry 5.0~\cite{chaccour2021seven} over wireless systems, such as 6G and beyond, there is a need for a fundamentally novel approach to performing communication in such systems. In particular, instead of relying on classical data-centric \ac{AI} techniques that limit wireless network generalizability and autonomy we must move towards \emph{generalizable, knowledge-based, and reasoning-driven} wireless networks~\cite{chaccour2022less}, through the concept of semantic communications. Semantic communications is a communication paradigm that involves transforming radio nodes into intelligent, reasoning agents that can unravel semantic content elements (meaning) and can communicate a \emph{minimal, efficient, and generalizable} representation that expresses the aforementioned meaning. In semantic communications the transmitter-receiver pair become a \emph{teacher-apprentice} that exchange a \emph{semantic language} rather than acting as a mere bit-pipeline.\\
\indent Hence, semantic communication is the path to \ac{AI}-native wireless networks~\cite{chaccour2022less}. Nonetheless, building a mature semantic language that can express structure, minimally and efficiently is a very challenging task. Low-layer source data tends to be very entangled and its structure is not directly interpretable as random information overlay its meaningful aspects. Thus, identifying the semantic content elements in a datastream and describing them via a \emph{minimal, efficient, and generalizable} representation becomes a very complex learning task and would lead to an overly complex semantic language. This complexity is not a result of the content complexity, but rather a consequence of attempting to learn the semantic content elements of possibly spurious random data points. It is thus necessary to devise a fully-fledged mechanism that enables disentangling \emph{semantic rich data} from \emph{spurious random information} in raw source data. This is a fundamental step in communicating a language that \emph{captures meaningful structure} and leads to a minimalist and efficient semantic communication system. Essentially, semantic communication systems cannot be robust, nor tend to generalizability, if their languages capture spurious information in the raw data.   
\subsection{Prior Works}
Recently, a number of works~\cite{xie2021deep, weng2021semantic, farshbafan2022curriculum, thomas2022neuro, xie2022task} studied the concept of semantic communication systems. In~\cite{xie2021deep}, a deep learning based semantic communication system for text transmission was devised. The work in~\cite{weng2021semantic} proposes deep learning approach to perform semantic communication for speech signals. The authors in~\cite{farshbafan2022curriculum} introduce a comprehensive semantic communications framework for enabling goal-oriented task execution. In~\cite{thomas2022neuro}, an emergent semantic communication system framework via a signaling game and a neuro-symbolic \ac{AI} approach is proposed. The authors in~\cite{xie2022task} explored task-oriented multi-user semantic communications to transmit data with single-modality and multiple modalities. While the works in~\cite{xie2021deep, weng2021semantic, farshbafan2022curriculum, thomas2022neuro, xie2022task} are interesting, they fail to acknowledge the entangled and intertwined nature of raw data. Furthermore, although the works in~\cite{xie2021deep, weng2021semantic, farshbafan2022curriculum, thomas2022neuro, xie2022task} study a language, yet they fail to distinguish the characteristics of a semantic language as well as the \ac{AI} and computational approaches needed to build a language from raw entangled data. Also, the works~\cite{xie2021deep, weng2021semantic, farshbafan2022curriculum, thomas2022neuro, xie2022task} do not take into account the fact that \emph{semantic-poor} data points are a complex learnable task, but a simple \emph{memorization} task, and thus are more efficiently transmitted classically. Finally, none of the works in~\cite{xie2021deep, weng2021semantic, farshbafan2022curriculum, thomas2022neuro, xie2022task} propose a fully fledged framework to disentangle raw data into meaningful semantic content elements that can be used to build a semantic language. Clearly, there is a gap in the research community to build semantic communication systems that can leverage raw data so as to build a language and transmit knowledge. 
\begin{figure*}
    \centering
		\includegraphics[width=0.59\textwidth]{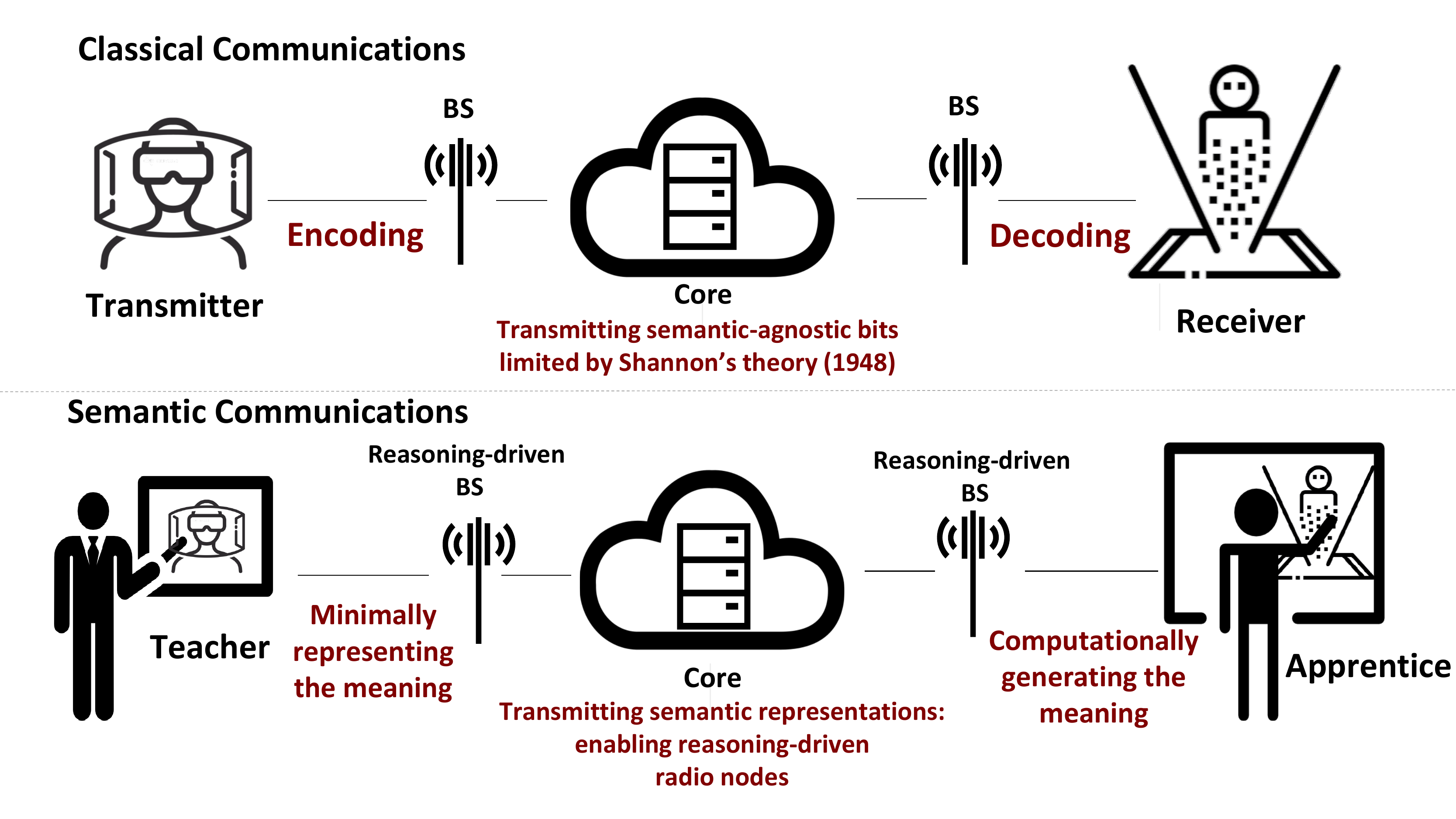}
		\caption{\small{Illustrative figure showcasing \ac{E2E} classical communication systems versus \ac{E2E} semantic communication systems}.}
    \label{fig:E2Esem}
    \vspace{-.35cm}
\end{figure*}
\subsection{Contributions}
The main contribution of this article is a novel contrastive learning approach to pre-process and disentangle raw data making it semantic-ready. In particular, a contrastive learning scheme is adopted at the transmitter side so as to disentangle raw source data into deep semantic clusters. This deep semantic clustering enables the separation of \emph{learnable data} from \emph{memorizable data} as well as data points belonging to different semantic content elements. In essence, \emph{learnable data} are data points with \emph{rich semantic information}, i.e., such data points can be expressed via a representation that has a meaningful semantic content element. Meanwhile, \emph{memorizable data points} are random data points that do not necessarily map to a particular significance, and are thus \emph{semantically poor}. Essentially, first, an instance discrimination contrastive learning task is performed whereby data points are compared to their neighbours at the instance-level. That is, instances undertake a perturbation which mimics a \emph{distribution shift}, then the teacher learns whether such data points are semantically close or not. Second, building on this instance discrimination task, a deep cluster discrimination task is performed so as to increase the cohesiveness between data points that belong to the same semantic content element, and disentangle semantic content elements of different meaning. The deep contrastive-semantic clusters built are further ranked per confidence level so as to label the least confident, as semantic poor data points, i.e., memorizable data that must be classically transmitted.  \emph{To the best of our knowledge this is the first work that pre-processes raw data via contrastive learning so as to disentangle semantic rich data points from semantic poor ones. This is also the first work that enables increasing the cohesiveness between semantically similar data points, and increases the disentanglement between semantically different raw data points.} Simulation results demonstrate that our contrastive learning disentanglement method enables realizing a large semantic impact ($\SI{71.9}{\%}$ higher than conventional semantic communication systems) despite increases in the content complexity. Our simulation results also show that the proposed contrastive learning approach yields representations of smaller length and thus enables a minimalist semantic communication system.
\section{System Model and Problem Formulation}
\subsection{Semantic Communication Model}
Consider a teacher $b$, who observes source information, which can stem from any use case (e.g. hologram, tactile feedback, image data) and is willing to transmit knowledge contained in the raw, low-layer, source datastream $\boldsymbol{X}$ to an apprentice $d$, as shown in Fig.~\ref{fig:E2Esem}. 
\begin{enumerate}
\item \emph{At the reasoning-driven transmitter side:} The goal of the teacher is to first \emph{disentangle} the semantic content elements $\boldsymbol{Y}=\{Y_1, \dots, Y_n\}$, ($n$ is the number of semantic content elements in the source information) contained in the datastream $\boldsymbol{X}$, i.e. separate the different meaning elements contained in the source data. Then, for every content element the teacher must learn a corresponding \emph{semantic representation} $\boldsymbol{Z}=\{Z_1, \dots, Z_n\}$ with desirable properties. Essentially, $\boldsymbol{Z}$ is the minimal, efficient, and generalizable description of $\boldsymbol{Y}$. 
\item \emph{At the reasoning-driven receiver side:} The goal of the apprentice is to \emph{understand} the meaning contained in the semantic representation $Z_i$ and use it to computationally generate semantic content elements $Y_i$. 
\end{enumerate}
To efficiently and effectively transmit semantic information between a teacher and an apprentice, it is necessary to devise a semantic language, that transforms the transmitter-receiver bit-pipeline into a structured information transfer. As we discussed in~\cite{chaccour2022less}, we can define a semantic language as follows:
\begin{definition}
A \emph{semantic language} $\mathcal{L}={(X_i, Z_i)},$ (or $\mathcal{L}={(X_{l,i}, Z_i)}$) is a dictionary (from a data structure perspective) that maps the source raw datastream (or an interventional learnable $X_{l,i}$) to their corresponding semantic representation $Z_i$, based on the identified semantic content elements $Y_i$.
\end{definition}
\subsection{Semantic Language Model}
In classical communications, the goal is to minimize the system entropy and maximize the channel capacity. In essence, statistical entropy characterizes the number of ``yes/no" questions we would have to ask in order to get complete information on the datastream we are dealing with. Entropy cannot be used in semantic communication systems as it solely depends on the freedom of choice of the transmitter rather than the meaning of the message. Thus, in a semantic communication system, the equivalent of ``entropy" is the ``language complexity" as discussed in~\cite{chaccour2022less}. Henceforth, in a semantic communication system, the goal of the complexity of the language is to characterize the difficulty of identifying and learning the semantic content elements in the raw datasteram $\boldsymbol{X}$. It is given by~\cite{chaccour2022less}:
\begin{equation}
    \Gamma(\mathcal{L})=\min_{p(\boldsymbol{Z}|\boldsymbol{X})}L_\mathcal{L}(p) + K(p),
\label{complexity}
\end{equation}
where, $L_\mathcal{L}(p)=\sum_{i=1}^N-\log p(Z_i|X_{i})$ is the cross-entropy loss, and $K(p)$ is the Kolmogorov complexity of the distribution $p(\boldsymbol{Z}|\boldsymbol{X})$.
From~\eqref{complexity}, one can observe:
\begin{itemize}
\item The language complexity, unlike Shannon's information-theoretic perspective, is a metric that depends on the meaning of representations. \eqref{complexity}  is a function of: a) the fidelity of the representation model in capturing the semantic content elements in the datastream, and b) the Kolmogorov complexity that characterizes the individuality of the semantic content elements as well as capturing the shortest effective binary description of $X_i$~\cite{chaccour2022less}. 
\item If the majority of the data lacks \emph{structure}, the \emph{learning task becomes increasingly difficult}. Nonetheless, this is not necessarily a result of complex content, but rather of a high number of random data points that lack semantic significance. In this setting, such data points must be \emph{memorized rather than learned}. That is, the semantic communication system must retain them as is and classically transmit them. 
\end{itemize}
 Furthermore, ~\eqref{complexity} is an ideal theoretical equation that measures complexity in terms of the idealized, shortest, and best obtainable model that can represent the semantic content elements in the data. The Kolmogorov complexity term in~\eqref{complexity} is not computable, i.e., there is no single function that will return the complexity of a particular representation binary string. Thus, for tractability and to capture a realistic representation model, based on fundamentals from transfer learning~\cite{achille2021information}, we can rewrite the language complexity as follows:
\begin{equation*}
    \Gamma(\mathcal{L},\zeta, \Lambda)=\mathbb{E}_{\theta~\Lambda(\theta|\mathcal{L})}\{L_\mathcal{L}(p_\theta(Z_i|X_i)\}+ \beta KL(\Lambda(\theta|\mathcal{L})||\zeta(\theta),
\end{equation*}
where the second term $KL(\Lambda(\theta|\mathcal{L})||\zeta(\theta)$ is the KL divergence that characterizes the information in the parameters of the representation model. Moreover, $\zeta$ and $\Lambda$ are, respectively, the pre-distribution and post-distribution of the representation model. Such distributions do not correspond to a prior or posterior of a Bayesian setting~\cite{achille2021information}. A pre-distribution is used to initialize the semantic representation model so as to initiate the gradual learning of a semantic language. Meanwhile the post-distribution is used to fine-tune the language acquired between the teacher and the apprentice. Both of these distributions depend on the level of symmetry between teacher and apprentice and their knowledge bases\footnote{Learning and optimizing such distributions is an important research avenue that is outside the scope of this paper.}. 
\subsection{Problem Formulation}
\indent As previously discussed, a language that lacks \emph{structure} will have a very large complexity due to the large amount of spurious random data points captured via representations. Remarkably, raw data tends to be very entangled, and it contains a large amount of random information. Hence, relying on the raw datastream $\boldsymbol{X}$ will lead to: a) an overly complex language with representations that cannot express meaningful semantic content elements, and b) such a problem mimicks the overfitting \ac{ML} scenario whereby the radio nodes have not learned representations but merely \emph{memorized} spurious patterns in the source information. To alleviate this problem, it is necessary to devise a comprehensive method that enables disentangling the learnable $\boldsymbol{X}_l$ data points from the memorizable ones $\boldsymbol{X}_m$. Performing such disentanglement technique will allow the radio node to:
\begin{itemize}
    \item Use $\boldsymbol{X}_l$ as their structure-rich datastream to build a semantic language, and learn efficient and minimal semantic representations. Thus, $\boldsymbol{X}_l$ will be transmitted semantically.
    \item Transmit $\boldsymbol{X}_m$ classically due to the lack of structure and rich semantic content elements in these data points. Attempting to perform reasoning or learning on $\boldsymbol{X}_m$ is a computationally inefficient scheme. Meanwhile, the classical communication scheme (which essentially targets characterizing the uncertainty in information) remains very efficient as it is not concerned with meaning.
    \end{itemize}
    \vspace{0.25cm}
The goal of this work is to propose a novel contrastive learning technique that enables pre-processing and disentangling raw source data, to ultimately separate learnable and memorizable data via instance and cluster discrimination. This approach also increases the cohesiveness between semantically similar data points and disentangling semantic different ones.
Next, we delve into the analytical foundation of our proposed contrastive learning framework.
\section{Contrastive Learning for Disentangling Semantic Content Elements}
Given a set of unlabeled\footnote{Unlabeled here refers to datastreams with unknown semantic content elements.} datastreams, we let $\kappa$ be the feature embedding network that extracts key structural information encoded in the low-dimensional vector subspace $\psi_\kappa: \boldsymbol{X} \rightarrow \boldsymbol{Z} \in \mathbb{R}^N$. Furthermore, we let $\phi_\varkappa$ be a classifying function that associates every semantic representation with its ground-truth semantic content element pseudo-label $\phi_\varkappa: \boldsymbol{Z} \rightarrow \boldsymbol{Y},$ where $\boldsymbol{Y} \in \mathcal{Y}$, $\mathcal{Y}$ is the universal set of possible semantic content elements that can be communicated. Essentially, the goal is that representations of the same cluster must share similar pseudo-labels vis-à-vis the semantic content elements they are representing. Next, we discuss how we perform instance discrimination to improve the generalizability of the representations that should be learned from the raw source data. Then, cluster discrimination is performed among the data points in $\boldsymbol{X}$ to optimize the boundaries between different semantic content elements and ultimately separate memorizable data. In essence, the higher confidence clusters are \emph{semantic rich} structures and are \emph{learnable} clusters, meanwhile lower confidence clusters are \emph{semantic poor} clusters with memorizable properties that must be classically transmitted. For instance, semantic-rich data can be thought as ``meaningful structural elements" of a hologram, meanwhile memorizable data may be random details in a hologram that do not have a meaning or structure (e.g. information that human beings would not notice about that hologram if trying to describe it). Furthermore, the instance discrimination technique performs this clustering while making sure that the mapping $\kappa$ leads to minimalist representations.
\subsection{Improving the Robustness of Semantic Representations via Instance Discrimination}
To perform instance discrimination~\cite{wu2018unsupervised}, we first consider every mapping function $\psi_\kappa$, and we construct another momentum encoder function\footnote{In this work, we consider the momentum contrast as our scheme for instantiation and instance discrimination.} $\psi_{\tilde{\kappa}}$ that shares an identical structure but independent parameters $\tilde{\kappa}$. Furthermore, for every raw datastream in $\boldsymbol{X}$, we randomly apply a set of transformations $\mathcal{X}$ so as to mimic future distribution changes, and improve the robustness of reasoning with respect to vertical generalizability (distribution shifts). For each instance of the datastream $X_i$, we represent two perturbed copies, namely, $\chi_1(X_i)$ and $\chi_2(X_i)$. As such, after passing such perturbed copies with the momentum encoder, we obtain $\boldsymbol{a}_i=\psi_\kappa(\chi_1(X_i))$ and $\boldsymbol{b}_i=\phi_\varkappa(\chi_2(X_i))$. In the instance discrimination, the contrastive learning task must match $\boldsymbol{a}_i$ and $\boldsymbol{b}_i$, against the contrastive set, composed by $K$ stale representations of the pseudo-negative samples of the semantic content element pseudo-labels~\cite{huang2021deep}:
% \begin{align}
% &\tilde{\boldsymbol{B}}_i=\{\tilde{\boldsymbol{b}}_1, \dots, \Tilde{\boldsymbol{b}_K}\} \nonumber\\
% &=\{\boldsymbol{\tilde{b}} | \boldsymbol{\tilde{b}}\in C_l  \hspace{0.2cm} \forall l \in \left[1, M\right] \text{and} \hspace{0.2cm} l \neq i \},
% \end{align}
\begin{align*}
&\tilde{\mathcal{B}}_i=\{\boldsymbol{\tilde{b}} | \boldsymbol{\tilde{b}}\in C_l  \hspace{0.2cm} \forall l \in \left[1, M\right] \text{and} \hspace{0.2cm} l \neq i \}=\{\tilde{\boldsymbol{b}}_1, \dots, \Tilde{\boldsymbol{b}_K}\},
\end{align*}
where $C_l$ is the memory bank used to manage the semantic content element and its respective cluster, and $M$ is the possible number of semantic content elements that might occur for the observed data stream.
The loss resulting from the instance discrimination step is given by~\cite{chen2020simple, hu2021region, huang2021deep}: 
\begin{equation}
    L_I(\boldsymbol{X}_i)=-\log\frac{\exp(\cos(\boldsymbol{a}_i, \boldsymbol{b}_i)/\tau)}{\sum_{\boldsymbol{\tilde{b}}\in \Tilde{\boldsymbol{B}}\cup\{\boldsymbol{b}_i\}} \exp(\cos(\boldsymbol{a}_i, \boldsymbol{\tilde{b}}))/\tau }.
\end{equation}
Next, we contrast random information with respect to clusters containing semantically similar content elements. 
\subsection{Disentangling Semantic Content Elements via Semantic Cluster Discrimination}
Essentially, data points that map to a single semantic content element (and its representation) must belong to a single semantic cluster free from any random information. Thus, in order to discover the random information existing within semantically similar clusters, we investigate the decision boundaries between such clusters. Then, we adopt an approach that enables closing the gap between semantically similar content elements, and eliminating random information from semantic rich clusters. In essence, cluster discrimination increases the compactness between intra-content data points, and increases the separation between inter-content data points, learnable, and memorizable data. Thus, given a sample $\boldsymbol{a_i}$, its probability of being in a semantically similar cluster is given by~\cite{chen2020simple, hu2021region, huang2021deep}: 
\begin{equation}
    P_{i,l}=\frac{\sum_{\boldsymbol{b}\in C_l}\exp(\cos(\boldsymbol{a}_i, \boldsymbol{\tilde{b}}))/\tau}{\sum_{l'=1}^N\sum_{\tilde{\boldsymbol{b}}\in C_{l'}}\exp({\cos(\boldsymbol{a}_i, \boldsymbol{\tilde{b}})/\tau)}},
\end{equation}
Then, we formulate the cluster discrimination loss as:
\begin{equation}
    L_D=\frac{1}{N}\sum_{i=1}^{N}\sum_{l=1}^{M}-P_{i,l} \log{P_{i,l}}.
\end{equation}
Thus, the total loss will be formulated as~\cite{chen2020simple, hu2021region, huang2021deep}: $L_T= \eta L_I+ \epsilon L_D$, where we can set $\eta=\epsilon=1$ in case we want to minimize the exhaustive per-dataset parameter tuning. To minimize this loss, the weights $\kappa$ and the decision boundaries of the semantic clusters are updated by back-propagation. Meanwhile, the momentum encoder $\tilde{\kappa}$ is updated via $\tilde{\kappa} \leftarrow  \omega \tilde{\kappa} +(1-\omega) \kappa$, where $\omega$ is the momentum encoder coefficient. 
To distinguish $\boldsymbol{X}_l$ from $\boldsymbol{X}_m$, the deep semantic clusters are ranked from highest confidence to the lowest confidence. Thus, the distinction is not binary: the higher the level of confidence, the more \emph{learnable and semantic rich} those data points are. Meanwhile, semantic deep clusters with the lowest level of confidence contain the data points that are the most \emph{memorizable and semantic poor}. Categorizing particular semantic deep clusters and their corresponding data points as $\boldsymbol{X}_m$ is a function of the lowest tolerable level of confidence. This is based on the tolerable language complexity in~\eqref{complexity}, the computation capability of the system, and the content complexity of the source data. 
\vspace{-.25cm}
% Next, we scrutinize our loss functions and prove that it converges to 
% \begin{lemma}
% When the cardinality of the contrastive set asymptotically increases and $M \rightarrow \infty$, for a fixed $\tau> 0$, the normalized contrastive loss converges to the following
%     \begin{align}
%     &\lim_{M \rightarrow \infty} (L_I+L_D)- \log{M} +\log{|\mathcal{Z}|} \nonumber \\ 
%     =-\frac{1}{\tau} 
%     % &=\mathbb{E}\left[H(p(.|\boldsymbol{Z}), q_h(.|\boldsymbol{Z})\right]
%     \end{align}
%     where $H$ is the entropy 
%     \begin{IEEEproof}
%     Proof was omitted due to space limitations
%     \end{IEEEproof}
%     \label{mylemma}
% \end{lemma}
% From Lemma~\ref{mylemma}, we can conclude that the language pre-processing, i.e.,  disentangling learnable and memorizable data points does not add substantial computational costs on the language complexity. Furthermore, given that the language now maps $\boldsymbol{X}_l$, instead of $\boldsymbol{X}$, the fidelity and minimalism of representations will significantly improve. Thus, owing to Lemma~\ref{mylemma}, one can see that the contrastive learning paradigm.
\section{Simulation Results and Analysis}
\subsection{Performance Evaluation via Semantic Key Performance Indicators (KPIs)}
Given that semantic communication systems are knowledge-centric, and thus leverage communication and computing resources; it is necessary to evaluate the performance of such systems using novel metrics. As such, in our simulation results, to evaluate the superiority of disentangling learnable and memorizable data via contrastive learning, we rely on the notion of \emph{semantic impact}. Essentially, as we have defined it in~\cite{chaccour2022less}, the semantic impact $\iota_{Y_i}$ measures the \emph{equivalent computational} packets generated via the transmission of a semantic representation. In other words, it measures the number of packets that a semantic representation can computationally generate, and thus highlights the amount of communication resources and time that could be saved when deploying semantic systems. 
\begin{figure}
    \centering
		\includegraphics[width=0.35\textwidth]{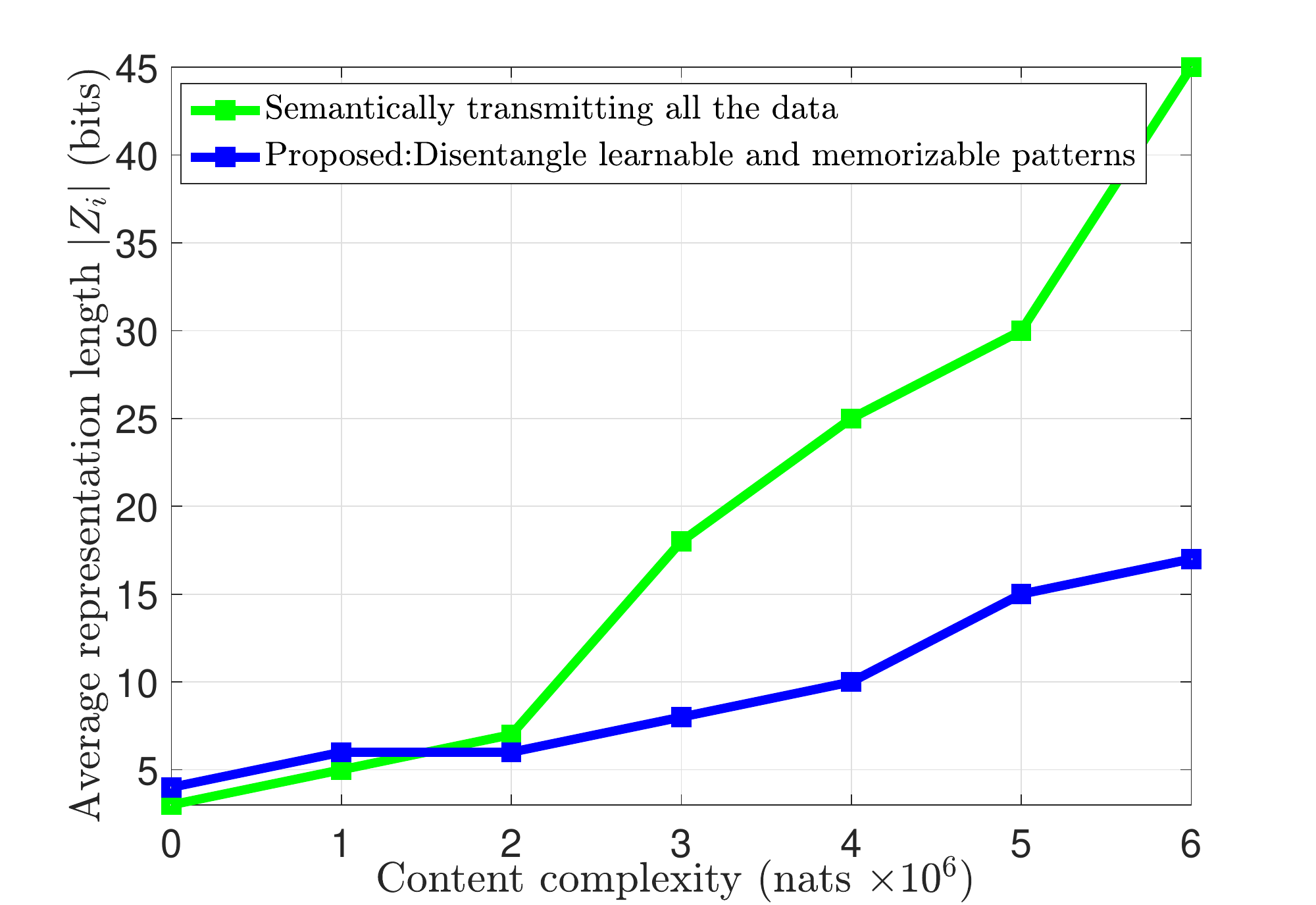}
		\caption{\small{Average representation length $|Z_i|$ (bits) versus content complexity (nats)}.}
    \label{fig:minimalism}
    \vspace{-.35cm}
\end{figure}
\begin{figure}
    \centering
		\includegraphics[width=0.35\textwidth]{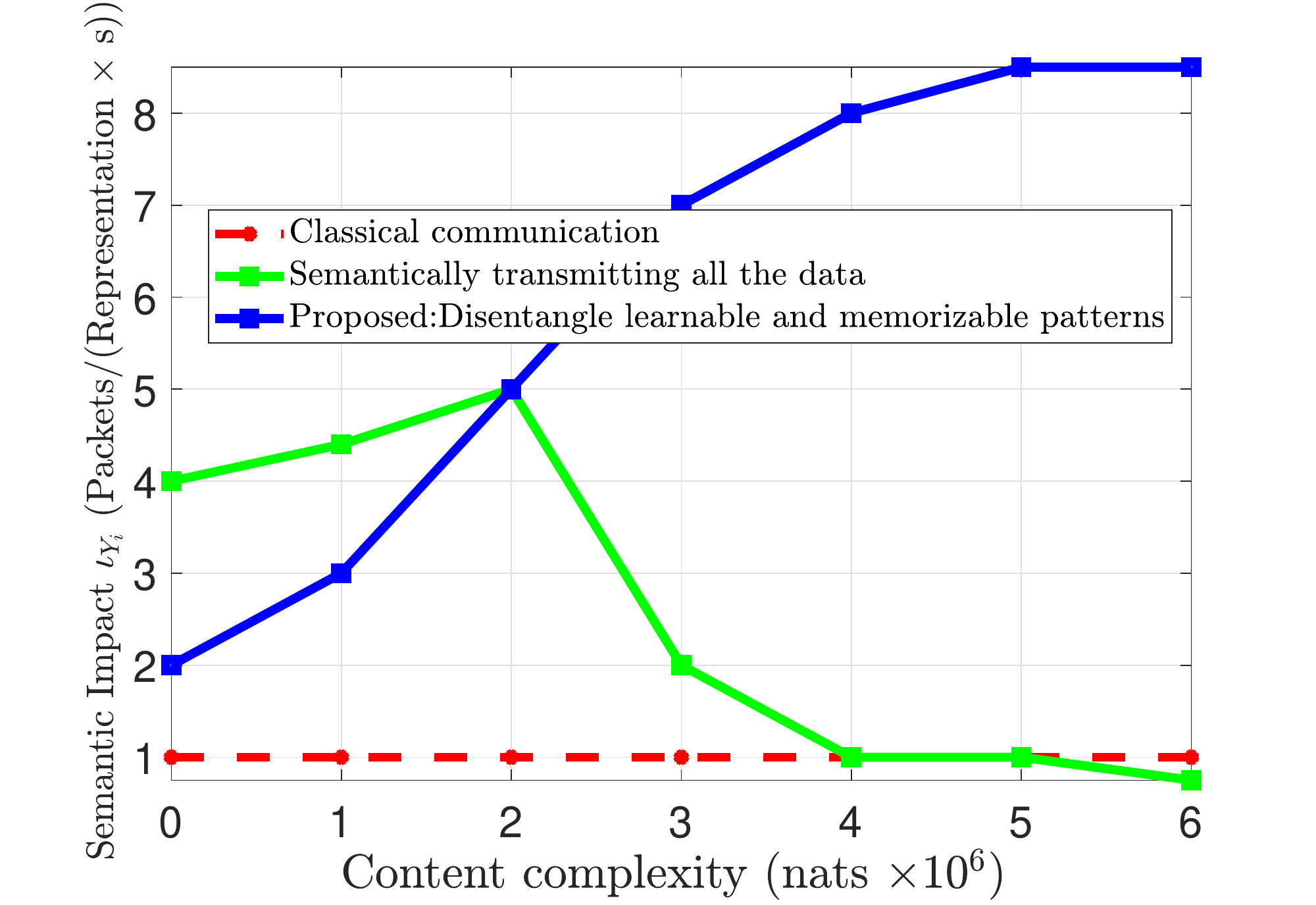}
		\caption{\small{Semantic impact $\iota_{Y_{i}}$ (Packets/(Representation $\times \SI{}{s} $)) versus content complexity (nats).}}
    \label{fig:semantic_impact}
        \vspace{-.35cm}
\end{figure}
\subsection{Simulation Results}
For our simulations, we consider $\boldsymbol{X}$ to be the low-level binary equivalent of a mixture model of CIFAR-10(/100)~\cite{krizhevsky2009learning} and ImageNet-10/Dogs~\cite{russakovsky2015imagenet}. The simulations adopted the stochastic gradient descent (SGD) optimizer, $\omega=0.9$, and $\tau=0.1$. The learning rate was set to $0.02$ across $\SI{200}{epochs}$ and a batch size of $32$~\cite{kandel2020effect}. \\
\indent In Fig.~\ref{fig:minimalism}, we can see that the average representation length increases with the content complexity. In fact, we can see that, for a low content complexity, semantically transmitting all the data might result in a smaller representation length. This is because the amount of random information, and thus $\boldsymbol{X}_m$ is considerably small. Meanwhile, as we increase the content complexity, we can see that semantically transmitting all the data is not a feasible approach anymore as the representation length steeply increases as we increase the content complexity. In contrast, when adopting our contrastive learning approach to pre-process the data, we can see that the average representation length is contained and minimalism can be achieved in the representation used, even for higher complexity orders. In fact, our representation is minimized by $57.22\%$ compared to the vanilla semantic approach. \\
\indent In Fig.~\ref{fig:semantic_impact}, we compare the semantic impact $\iota_{Y_i}$ for classical communications, fully transmitting all data via semantic communications, and adopting our proposed contrastive learning approach to disentangle $\boldsymbol{X}_l$ and $\boldsymbol{X}_m$. We can see that while for a low content complexity, semantically transmitting all the data might be a feasible solution, however, as we increase the content complexity we can see how the semantic impact drastically decreases. Meanwhile, when contrastive learning is adopted to pre-process the language, we can see that the semantic impact grows despite increases in the content complexity. In fact, a $71.9\%$ improvement is achieved compared to a semantic communication system that lacks language pre-processing.
% In fact, in Shannon's dimension, the entropy was a proper metric to quantify the complexity of a random variable containing information. This is due to the fact that the overall process relied on tasks like compression and transmission which essentially require reconstructing the original random variable. Meanwhile, when reasoning is added to the equation, the whole premise of information theory that relies on entropy needs to be revisited and re-engineered so as to characterize reasoning-driven communications.\\
%
%
%To understand the necessary measures to characterize semantic communications, we need to dwell on the fundamentals of classical information theory first. In essence, entropy in classical communications is used to measure the amount of missing information before reception. In other words, in the case transmitted messages are equiprobable, the Shannon entropy is basically the number of binary questions needed to determine the content of the message. Here, a ``bit" is a micro-state affecting the overall macro-state of the system. 
\section{Conclusion}\label{Sec:Conclusion}
In this paper, we have proposed a novel contrastive learning approach to pre-process and disentangle the raw data used in semantic communication systems. In particular, our contrastive learning approach performs instance and cluster discrimination on raw data points. In essence, we increase the cohesiveness between data points that belong semantically similar content elements and disentangle data points belonging to semantically different content elements. Subsequently, we rank the deep semantic clusters and consider the least confident ones as memorizable, semantic-poor data. Simulation results demonstrate the superiority of our contrastive learning approach compared to pure semantic and classical communications vis-à-vis semantic impact and minmalism of the representation.
\bibliographystyle{IEEEtran}
\def\baselinestretch{0.88}
\bibliography{bibliography}
\end{document}

%% file: acronyms.tex
\glsdisablehyper
\acro{D2D}{device-to-device}
\acro{SIR}{signal-to-interference-ratio}
\acro{SINR}{signal-to-interference-plus-noise-ratio}
\acro{PCP}{Poisson cluster process}
\acro{CoMP}{coordinated multi-point}
\acro{BS}{base station} 
\acro{MD-CoMP}{macrodiversity CoMP transmission}
\acro{MAC}{medium-access-control}
\acro{JT-CoMP}{joint transmission CoMP}
\acro{CoMP-JT}{coordinated multipoint joint transmission}
\acro{SBS}{small base station}
\acro{MDSD}{multiple devices to a single device}
\acro{MDS}{maximum distance separable}
\acro{SCN}{small cell network}
\acro{PPP}{Poisson point process}
\acro{TCP}{Thomas cluster process}
\acro{CSI}{channel state information}
\acro{PDF}{probability distribution function}
\acro{PMF}{probability mass function}
\acro{RV}{random variable}
\acro{i.i.d.}{independently and identically distributed}
\acro{MBMS}{multimedia broadcasting multicasting service}
\acro{EE}{energy efficiency}
\acro{HCP}{hard-core placement}
\acro{CCDF}{complementary cumulative distribution function}
\acro{CDF}{cumulative distribution function}
\acro{PC}{probabilistic caching}
\acro{RC}{random caching}
\acro{CPF}{caching popular files} 
\acro{PGFL}{probability generating functional}
\acro{KKT}{Karush-Kuhn-Tucker}
\acro{PGF}{point generating function}
\acro{SCA}{successive convex approximation}
\acro{HD}{high-definition}
\acro{FHD}{full-high-definition}
\acro{UHD}{ultra-high-definition}
\acro{VR}{virtual reality}
\acro{AR}{augmented reality}
\acro{5G}{fifth-generation}
\acro{QoS}{quality-of-service}
\acro{QoE}{quality-of-experience}
\acro{IoT}{internet of things}
\acro{MHCPP}{Matern hardcore point process}
\acro{LoS}{line-of-sight}
\acro{NLoS}{non-line-of-sight}
\acro{PSD}{power spectral density}
\acro{MEC}{mobile edge computing}
\acro{E2E}{end-to-end}
\acro{THz}{terahertz}
\acro{CLT}{central limit theorem}
\acro{HQ}{High Quality}
\acro{eMBB}{enhanced mobile broadband}
\acro{URLLC}{ultra reliable low latency communications}
\acro{mmWave}{millimeter wave}
\acro{EVT}{extreme value theory}
\acro{GEV}{generalized extreme value}
\acro{HR2LLC}{high-rate and high-reliability low latency communications}
\acro{HF}{high frequency}
\acro{TVaR}{tail-value-at-risk}
\acro{UE}{user equipment}
\acro{RIS}{reconfigurable intelligent surfaces}
\acro{MIMO}{multiple-input and multiple-output}
\acro{HMD}{head-mounted display}
\acro{AI}{artificial intelligence}
\acro{ML}{machine learning}
\acro{HITRAN}{high resolution transmission}
\acro{OFDM}{orthogonal frequency division multiplexing}
\acro{PAPR}{peak-to-average power ratio}
\acro{MR}{mixed reality}
\acro{EVaR}{entropic value-at-risk}
\acro{DNN}{deep neural network}
\acro{MDP}{Markov decision process}
\acro{RL}{reinforcement learning}
\acro{RNN}{recurrent neural network}
\acro{ANN}{artificial neural networks}
\acro{LSTM}{long short-term memory}
\acro{ReLu}{rectified linear unit}
\acro{VaR}{value-at-risk}
\acro{SNR}{signal-to-noise ratio}
\acro{RF}{radio frequency}
\acro{CVaR}{conditional value-at-risk}
\acro{AVaR}{average value-at-risk}
\acro{ES}{expected shortfall}
\acro{AoI}{age of information}
\acro{PAoI}{peak age of information}
\acro{QoPE}{quality of physical experience}
\acro{LCFS}{last come first served}
\acro{FCFS}{first come first served}
\acro{SC-FDM}{single carrier frequency-division multiplexing}
\acro{ToA}{time of arrival}
\acro{MUSIC}{multiple signal classification}
\acro{IoE}{Internet of Everything}
\acro{6DoF}{six degrees of freedom}
\acro{OFDMA}{orthogonal frequency-division multiple access}
\acro{AoSA}{array of subarray}
\acro{XR}{extended reality}
\acro{AoA}{angle of arrival}
\acro{ULA}{uniform linear array}
\acro{AoD}{angle of departure}
\acro{EM}{electromagnetic}
\acro{IoI}{Internet of Intelligence}
\acro{MISO}{multiple-input and single-output}
\acro{CRAS}{connected robotics and autonomous systems}
\acro{FL}{federated learning}
\acro{GAN}{generative adversarial network}
\acro{mMTC}{massive machine type communications}
\acro{Tbps}{terabit per second}
\acro{NTN}{Non-Terrestrial Network}
\acro{SLAM}{simultaneous localization and mapping}
\acro{IS}{information shower}
\acro{OAM}{orbital angular momentum}
\acro{NOMA}{non-orthogonal mutliple access}
\acro{CPS}{cyber physical system}
\acro{IoNT}{Internet of Nano-Things}
\acro{TDD}{time division duplex}
\acro{CDMA}{code division multiple access}
\acro{OMA}{orthogonal multiple access}
\acro{VLC}{visible light communications}
\acro{UAV}{unmanned aerial vehicle}
\acro{LEO}{low earth orbit}
\acro{DL}{deep learning}
\acro{KPI}{key performance indicator}
\acro{NLP}{natural language processing}
\acro{SCM}{structural causal model}
\acro{RAN}{radio access network}